\documentclass{article}
\usepackage{ijcai24}

\usepackage{times}
\usepackage{soul}
\usepackage{url}
\usepackage[hidelinks]{hyperref}
\usepackage[utf8]{inputenc}
\usepackage[small]{caption}
\usepackage{graphicx}
\usepackage{amsmath}
\usepackage{amsthm}
\usepackage{booktabs}
\usepackage{algorithm}
\usepackage{algorithmic}
\usepackage[switch]{lineno}

\usepackage{pifont}
\usepackage{multicol}
\usepackage{multirow}

\usepackage{xcolor}

\usepackage{cleveref}
\crefname{figure}{Figure}{Figures} 
\crefname{table}{Table}{Tables}   
\crefname{section}{Section}{Sections} 

\title{Bridging Generative and Discriminative Learning: \\ Few-Shot Relation Extraction via Two-Stage Knowledge-Guided Pre-training}
\author{Quanjiang Guo$^{1}$, Jinchuan Zhang$^{1}$, Sijie Wang$^{2}$, Ling Tian$^{1}$, Zhao Kang$^{1}$\thanks{Corresponding Author},\\ Bin Yan$^{3}$, Weidong Xiao$^{4}$\\
        $^1$ University of Electronic Science and Technology of China, Chengdu, China \\ 
        $^{2}$ Nanyang Technological University, Singapore  \\
        $^{3}$ Information Engineering University, Zhengzhou, China  \\
        $^{4}$ National University of Defense Technology, Changsha, China  \\
        guochance1999@163.com, jinchuanz@std.uestc.edu.cn,
        wang1679@e.ntu.edu.sg,\\
        \{lingtian, zkang\}@uestc.edu.cn,  tom.yan@gmail.com, wdxiao@nudt.edu.cn}

\begin{document}

\maketitle

\begin{abstract}

Few-Shot Relation Extraction (FSRE) remains a challenging task due to the scarcity of annotated data and the limited generalization capabilities of existing models. Although large language models (LLMs) have demonstrated potential in FSRE through in-context learning (ICL), their general-purpose training objectives often result in suboptimal performance for task-specific relation extraction. To overcome these challenges, we propose TKRE (Two-Stage Knowledge-Guided Pre-training for Relation Extraction), a novel framework that synergistically integrates LLMs with traditional relation extraction models, bridging generative and discriminative learning paradigms. TKRE introduces two key innovations: (1) leveraging LLMs to generate explanation-driven knowledge and schema-constrained synthetic data, addressing the issue of data scarcity; and (2) a two-stage pre-training strategy combining Masked Span Language Modeling (MSLM) and Span-Level Contrastive Learning (SCL) to enhance relational reasoning and generalization. Together, these components enable TKRE to effectively tackle FSRE tasks. Comprehensive experiments on benchmark datasets demonstrate the efficacy of TKRE, achieving new state-of-the-art performance in FSRE and underscoring its potential for broader application in low-resource scenarios. \footnote{The code and data are released on 
\url{https://github.com/UESTC-GQJ/TKRE}.}

\end{abstract}
%



\vspace{-0.5cm}
\section{Introduction}
\begin{figure}[ht]
    \small
	\centering
	\includegraphics[width=1\linewidth]{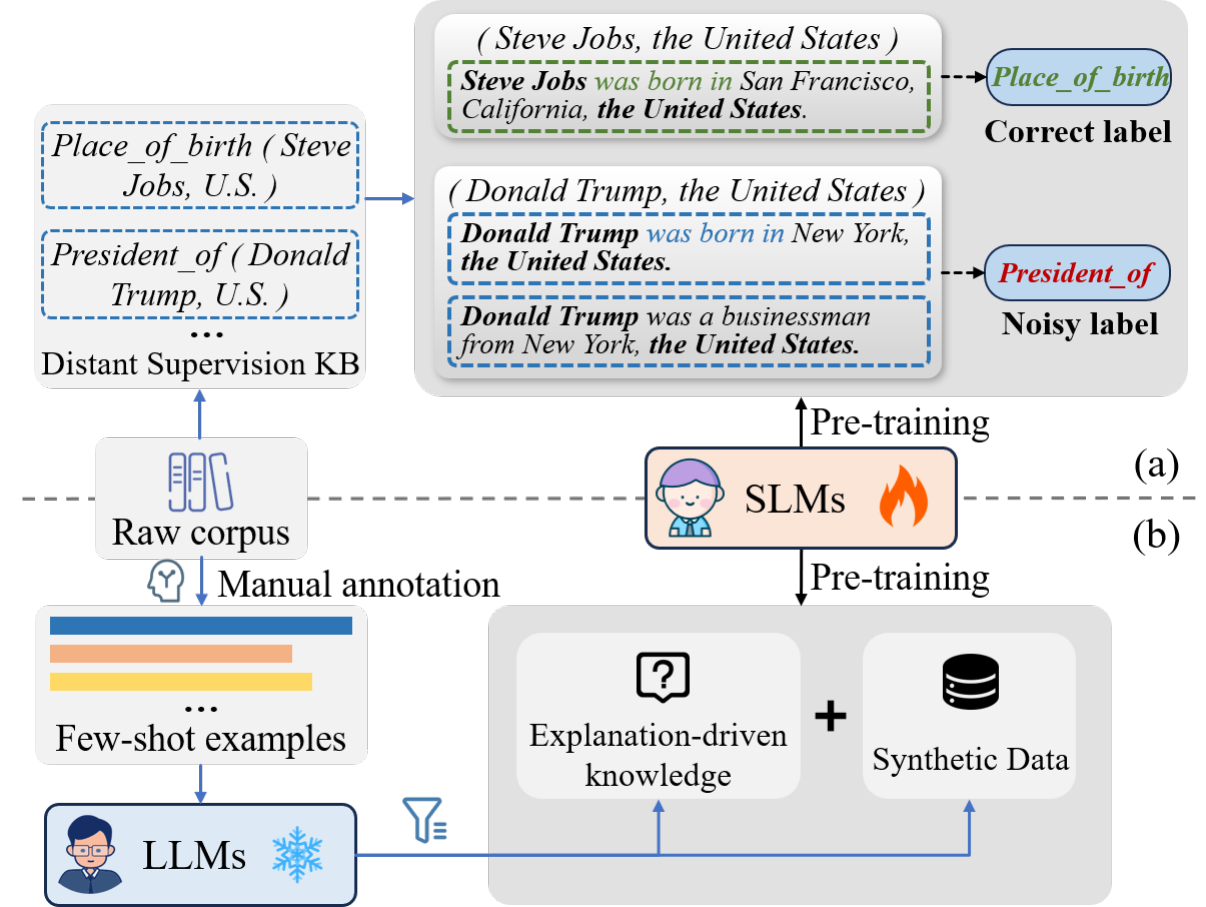}
	\caption{(a) Distant supervision is an automated and efficient approach for generating labeled data, but its effectiveness depends on the quality of the knowledge base. Challenges such as noisy labels and context loss are common issues with distant supervision. (b) In contrast, our approach leverages explanation-driven knowledge and synthetic data automatically generated by LLMs, which significantly enhance the performance of small language models (SLMs).}
	\label{intro1}
\end{figure}

Relation Extraction (RE) focuses on identifying semantic relationships between entities in unstructured text, serving as a cornerstone for knowledge graph construction and event analysis. Although pre-trained language models (PLMs)~\cite{wei2020novel,li2022fastre,wang2023pascore,zhu2024fcds} have significantly advanced RE performance, their success hinges on the availability of large-scale annotated datasets—a requirement that is often unattainable in real-world applications such as rare disease relationship mining and financial event detection. These scenarios are particularly challenging due to the severe scarcity of annotated examples. As a result, Few-Shot Relation Extraction (FSRE) has emerged as a critical research area, necessitating the development of models capable of generalizing from only a handful of labeled examples per relation type.

Existing FSRE methodologies face three critical challenges.
\textbf{Data Scarcity}: Traditional approaches rely on extensive annotations to distinguish between relations. While distant supervision~\cite{vrandevcic2014wikidata} generates labeled data by aligning with knowledge bases, it introduces noisy labels and domain mismatches~\cite{soares2019matching,peng2020learning,qin2021erica}, as shown in Figure \ref{intro1}(a).
\textbf{Limited Generalization}: Models trained on in-distribution data struggle to adapt to cross-domain or fine-grained relations. For instance, discriminative models frequently misclassify semantically similar relations, such as confusing ``Place\_of\_birth" with ``Place\_of\_death", with error rates exceeding 30\%~\cite{alt2020tacred}.
\textbf{Generative-Discriminative Disconnect}: Large language models (LLMs) like GPT-4~\cite{achiam2023gpt} show potential for generating synthetic data but treat RE as a black-box text generation task~\cite{li2023synthetic,xu2023unleash}. This results in a misalignment between the outputs of LLMs and the structured learning objectives required by FSRE models.

To address these challenges, we propose Two-stage Knowledge-guided Relational Extraction (TKRE), a novel framework that integrates generative and discriminative paradigms through structured knowledge infusion. In the first stage, we leverage large language models (LLMs) to generate relational logic explanations (e.g., ``Subject X exhibits relation R with Object Y because ..."), explicitly capturing fine-grained semantic dependencies between entities. These explanations create interpretable reasoning pathways that guide downstream model training. In the second stage, we employ constrained decoding strategies to generate schema-compliant training instances by enforcing entity type compatibility and relation-specific lexico-syntactic patterns. This dual approach ensures that the synthetic data preserve domain-specific relational semantics while mitigating distributional shifts, as illustrated in Figure \ref{intro1}(b).

Stage 2 introduces a two-stage knowledge-guided pre-training strategy that combines Masked Span Language Modeling (MSLM) and Span-level Contrastive Learning (SCL). MSLM focuses on selectively masking relation-indicative spans and optimizing their reconstruction, enhancing the model's ability to capture contextual dependencies and infer relational semantics. Complementing this, SCL operates within a contrastive learning framework, where positive spans (semantically aligned with target relations) are contrasted with negative spans (relationally discordant but contextually plausible). This approach sharpens the model’s discriminative boundaries for entity interaction modeling, fostering a more precise understanding of relational structures.

\begin{figure*}[ht]
    \small
	\centering
	\includegraphics[width=0.99\linewidth]{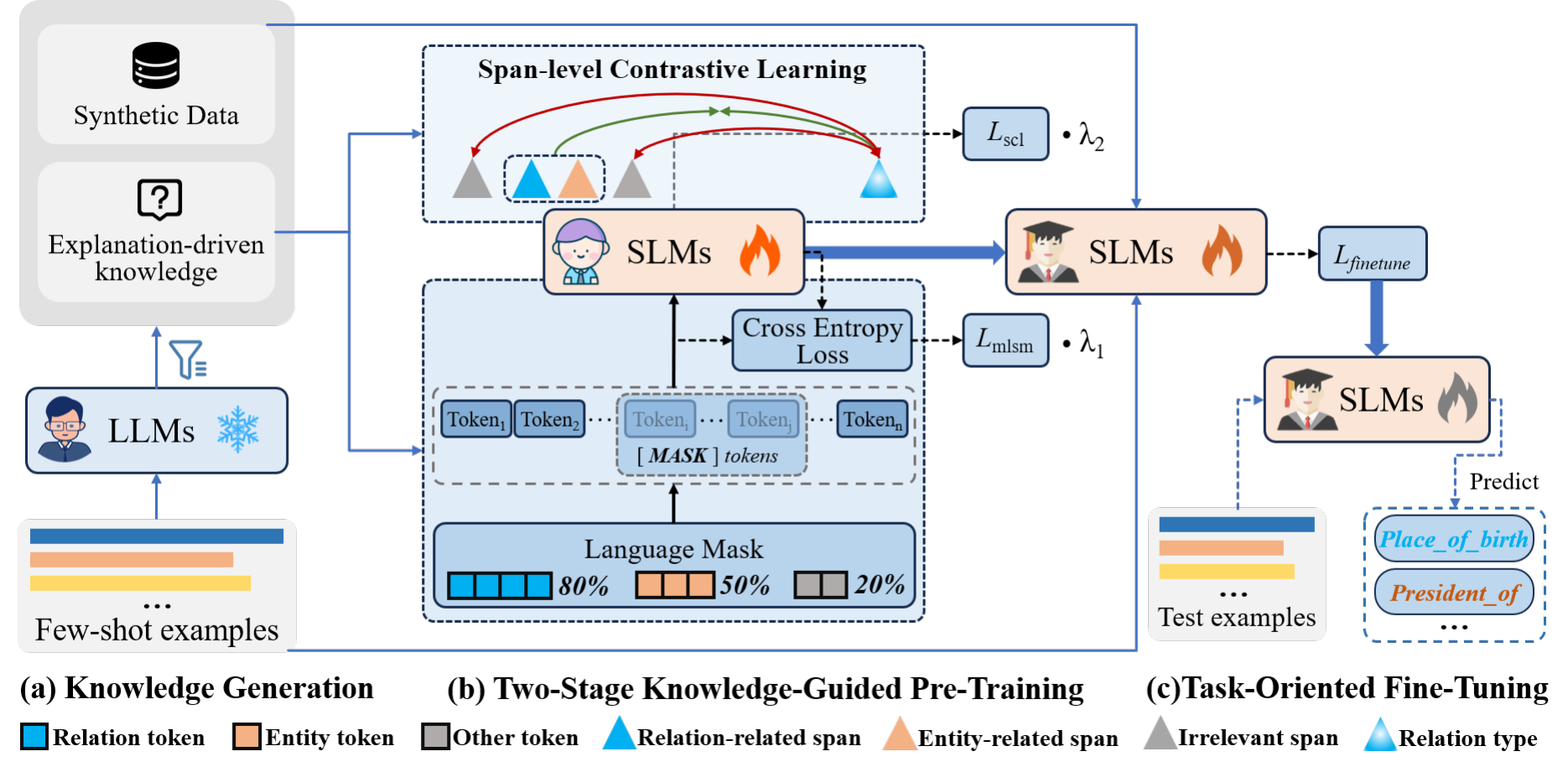}
	\caption{Overall framework of the proposed \textsc{TKRE}. (a) Leveraging LLMs to generate explanation-driven knowledge and schema-constrained synthetic data. (b) Implementing a two-stage knowledge-guided pre-training framework to enhance the model’s ability to understand relational structures and contextual dependencies. (c) Performing task-oriented fine-tuning and evaluation on the test dataset to validate the model’s effectiveness. }
	\label{tkre}
\end{figure*}

As illustrated in \cref{tkre}, TKRE establishes a self-reinforcing knowledge integration cycle: (a) relational explanations guide the generation of synthetic data, ensuring semantic consistency; (b) the synthetic data is leveraged during pre-training to incorporate generative knowledge; and (c) pre-training objectives explicitly align the representations of synthetic and golden data. This closed-loop architecture facilitates the iterative refinement of relational understanding through knowledge distillation and representation alignment, effectively bridging the gap between generative and discriminative paradigms.

In summary, our contributions are as follows:
\begin{itemize}
    \item We leverage LLMs to automatically generate explanation-driven knowledge and structured synthetic data, addressing data scarcity while enhancing the traditional RE model's comprehension of underlying relational logic.
    \item We introduce a novel two-stage knowledge-guided pre-training framework that seamlessly integrates generative knowledge with discriminative objectives, enabling models to effectively learn from sparse annotations and noisy synthetic data simultaneously.
    \item Extensive experiments on four benchmark datasets demonstrate that TKRE significantly improves the performance of traditional FSRE models, achieving F1 gains of 7.8\% and 5.0\% over strong baselines (TYP Marker, GenPT) and outperforming both pure LLM-based and hybrid methods in few-shot settings.
\end{itemize}

\section{Related Work}

\paragraph{Few-Shot Relation Extraction with Traditional Models.} Traditional FSRE methods predominantly rely on discriminative learning to optimize $P(y|x)$, but they face significant challenges under low-resource conditions. Approaches such as ~\cite{peng2020learning}, ~\cite{zhou2022improved} and ~\cite{zhang2024tpn} enhance intra-class representation consistency but struggle to generalize to fine-grained relations, due to their reliance on rigid in-distribution assumptions. Prompt-based methods such as~\cite{han2022generative} and PTRE~\cite{zhang2024prompt}) reformulate classification as an infilling task, yet require precise semantic alignment and fail to scale with extreme data scarcity. Distant supervision~\cite{mintz2009distant} alleviates annotation costs via knowledge base alignment, and many recent works investigate how to improve learning with such distant data~\cite{soares2019matching,peng2020learning,qin2021erica}. Despite its benefits, distant supervision introduces noisy labels and domain mismatches, which significantly undermine model robustness.

These limitations arise from a common bottleneck: \textit{discriminative models excel at distinguishing predefined classes but lack the ability to reason over unseen relational logic or contextual dependencies}. This deficiency is particularly critical in FSRE, where success depends on capturing subtle semantic distinctions and adapting to cross-domain variations. To address these challenges, our proposed TKRE framework integrates generative knowledge to bridge this gap, enabling more robust relational reasoning and improved adaptability.

\paragraph{Large Language Models for FSRE.} The emergence of LLMs like GPT-4~\cite{achiam2023gpt} and LLaMA-2~\cite{touvron2023llama} has generated significant interest in leveraging their \textbf{in-context learning} capabilities for FSRE. Early efforts~\cite{xu2023s2ynre,li2023synthetic,xu2023unleash} query LLMs using task descriptions and demonstrations, achieving only marginal improvements over traditional baselines. These approaches, however, treat LLMs as black-box generators and fail to align their \textbf{generative knowledge} (e.g., contextual reasoning and dependency modeling) with the structured discriminative objectives of RE models. As a result, the generated outputs often lack task-specific structural awareness, leading to suboptimal performance on fine-grained relations~\cite{li2023synthetic}. Furthermore, directly fine-tuning LLMs is computationally prohibitive~\cite{yang2024harnessing}, and their zero-shot performance is constrained by the complexity of prompt design~\cite{whitehouse2023llm,guo2025baner}. These limitations underscore the urgent need for frameworks that effectively bridge generative and discriminative learning paradigms to unlock the full potential of LLMs for FSRE.

\paragraph{Hybrid Methods: Combining LLMs with Traditional Models.} Recent hybrid approaches attempt to address data scarcity by leveraging LLMs to augment training data for smaller RE models. Editing-based methods~\cite{wei2019eda,jiao2020tinybert} replace tokens with synonyms or back-translated sentences but often fail to maintain relational consistency. Generative methods~\cite{anaby2020not,schick2021exploiting} use LLM prompting to synthesize labeled instances, yet their outputs frequently suffer from noise and lack diversity in structured RE tasks. Semi-supervised techniques~\cite{vu2021strata} combine synthetic and golden data using confidence filtering or auxiliary tasks but treat synthetic data as static inputs rather than dynamic knowledge carriers. For instance, ~\cite{xu2023s2ynre} introduces a two-stage self-training framework but does not explicitly model entity-span interactions. Similarly, ~\cite{wang2023instructuie} generates label-conditioned data but neglects alignment between synthetic examples and relational reasoning. These methods fail to fully exploit the potential synergy between generative and discriminative paradigms, limiting their ability to transfer LLM-derived knowledge into task-specific representations effectively.


\section{Methodology}
In this section, we present the detailed modules of our TKRE: explanation-driven knowledge generation, two-stage knowledge-guided pre-training, and task-oriented fine-tuning. Figure \ref{tkre} depicts the overall framework.

\subsection{Task Formulation.} Let $\mathcal{R} = \{r_1, r_2, \dots, r_{|\mathcal{R}|}\}$ represent the set of $|\mathcal{R}|$ relation types in the dataset (e.g., TACRED contains $ |\mathcal{R}| = 42 $). Given a sentence $x$ with two marked entities $ e_{\mathrm{sub}} $ and $ e_{\mathrm{obj}} $, the goal is to predict the relation $ r \in \mathcal{R} $ between these entities. In the few-shot setting, only $ K $ labeled examples $ \mathcal{D}_{\mathrm{train}} = \{ ( x, e_{\mathrm{sub}}, e_{\mathrm{obj}}, r )_i \}_{i=1}^{K \times |\mathcal{R}|} $ per relation type are provided for training, where $ K $ is a small constant (e.g., $ K = 8 $).

\subsection{Data Preparation}
To address the challenges of data scarcity and limited generalization, we employ LLMs to generate \textbf{knowledge explanations} and \textbf{synthetic examples} for each relation type, followed by multi-task pre-training and fine-tuning.

\subsubsection{Explanation-Driven Knowledge Generation}

To further improve domain adaptation and enhance the task relevance of the pre-training strategy, we utilize LLMs to construct task-oriented, generated knowledge. LLMs, trained on diverse corpora that likely encompass domains relevant to NER tasks, provide a rich source of contextual and domain-specific information. However, directly fine-tuning LLMs is computationally expensive and resource-intensive, making it impractical for many downstream tasks.

\begin{figure}[h]
    \small
	\centering
	\includegraphics[width=0.7\linewidth]{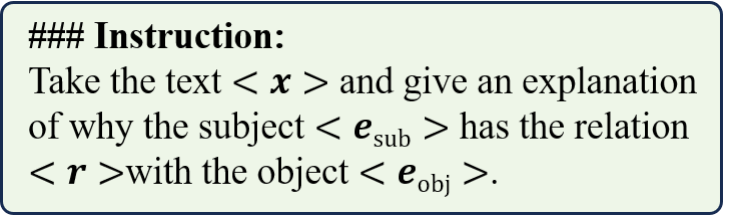}
        \caption{Instruction of explanation-driven knowledge generation.}
	\label{method1}
\end{figure}

To address this, we design an intuitive instruction to guide the LLM in explaining why a given text span should be recognized as an entity, thereby generating a task-oriented corpus. For a sentence $x \in \mathcal{D}_{\mathrm{train}}$, the golden relation type $r$, and an entity pair ($e_{\mathrm{sub}}$, $e_{\mathrm{obj}}$) $\in$ $x$, the instruction $X$ is constructed as described in \cref{method1}.

\vspace{-0.2cm}
Given this instruction $X$, we denote LLM as $f_\mathrm{LLM}$. The generated sequence regarding the explanation of relation $< y >$ is denoted as 
$Y=f_\mathrm{LLM}(X)=\{y_i\}^{L}_{i=1}$, where $L$ is defined as the length of the generated sequence, i.e., the number of tokens in $Y$. For the classic auto-regressive generative model, the generated sequence $Y$ is predicted by the following conditional probability:
\vspace{-0.2cm}
\begin{equation}
    {p}(Y \mid X) = \prod_{t=1}^{L} {p}(y_t \mid X, y_{<t}),
\end{equation}
where $y_{<t}$ represents the tokens before $y_t$.

Consequently, we can obtain several sentences of a relation extraction flow by reasoning in the raw textual context $<x>$, such as the grey part in Figure ~\ref{intro1}. Then, with respect to all relation types in $\mathcal{D}_{\mathrm{train}}$, we employ the frozen LLM $\mathcal{M}$ to obtain $K$ explanations cluster of each relation type. Formally,

\begin{equation}
    \{ \mathbf{Y}_{i} \} = \mathcal{M}_{\mathrm{Frozen}}(X_{r_i}), r_i \in \mathcal{R}
\end{equation}

where $X_{r_i}$ denotes the instruction $X$ with the corresponding slots of relation $r_i$. Following \cite{qin2021erica}, we build the explanation-driven knowledge corpus $\mathcal{C}$ from the labeled raw texts in $\mathcal{D}_{\mathrm{train}}$.

\subsubsection{Schema-Constrained Data Generation}

\begin{figure}[h]
    \small
	\centering
	\includegraphics[width=0.7\linewidth]{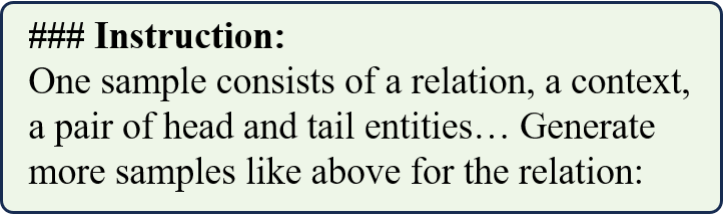}
        \caption{Instruction of schema-constrained synthetic data generation.}
	\label{method3}
\end{figure}

In this section, we aim to perform data augmentation using LLMs to enrich the training data for RE. Inspired by \cite{xu2023unleash}, we design prompts that inform the LLM about the essential components of an RE training sample: context text, subject entity, object entity, subject entity type, object entity type, and the relation. The LLM is then guided to generate additional pseudo RE samples based on this structure. To ensure the augmented data aligns with the desired format, we apply predefined rules, such as regular expressions, to process and transform the LLM-generated outputs. The specific instruction provided to the LLM is illustrated in Figure \ref{method3}.
\vspace{-0.5cm}
\subsection{Two-Stage Knowledge-guided Pre-training}

In this section, we present a \textit{two-stage knowledge-guided pre-training framework} that integrates \textbf{Masked Span Language Modeling (MSLM)} and \textbf{Span-level Contrastive Learning} tasks. This framework is designed to equip traditional FSRE models with the ability to effectively capture contextual relationships between entities while simultaneously learning robust span-level representations. By combining these complementary tasks, the pre-training strategy enhances the model’s capability to reason over entity relationships and adapt to fine-grained relational distinctions.

\subsubsection{Masked Span Language Modeling}

MSLM is an extension of traditional Masked Language Modeling (MLM) \cite{devlin2019bert}, where instead of masking individual tokens, we mask entire spans (sequences of tokens). Inspired by the better learning ability of span masking \cite{liu2021} and recent advances in span-based masking methods \cite{zhang2024cross}, we adopt this strategy to improve the model’s ability to capture long-range dependencies and complex relations between entities.

Let $ x = \langle x_1, x_2, \dots, x_n \rangle $ represent a given sentence, where $ x_i \in \mathcal{C} $ is a token from the explanation-driven knowledge corpus $\mathcal{C}$, and $ n $ is the sentence length. In MSLM, we mask a continuous span $ \mathcal{S} = \langle x_i, x_{i+1}, \dots, x_j \rangle $ by replacing it with a special sentinel token, such as $[MASK]$. The task is to predict the masked span $ \mathcal{S} $ based on the surrounding context.

Given a masking strategy with different probabilities for different types of spans:
\begin{itemize}
    \item 80\% probability to mask relation spans,
    \item 50\% probability to mask subject/object entity spans,
    \item 20\% probability to mask other words.
\end{itemize}

We define a custom probability distribution for masking tokens based on their linguistic role in the sentence. The mask distribution \( M = \langle m_1, m_2, \dots, m_n \rangle \) is determined by the specific type of each token \( x_i \). For each token \( x_i \), the masking probability \( p_i \) is defined according to the following rules:

\begin{equation}
p_i = 
\begin{cases}
    0.8 & \text{if } x_i \text{ belongs to relation span,} \\
    0.5 & \text{if } x_i \text{ belongs to subject/object entity span,} \\
    0.2 & \text{if } x_i \text{ belongs to other type of words.} \\
\end{cases}
\end{equation}

This strategy allows for a more targeted masking process, where different parts of the sentence are masked with different probabilities based on their syntactic roles. We model this process using a Bernoulli distribution for each token, where the probability of masking a token \( x_i \) is determined by its type:

\begin{equation}
P(M = m_i | p_i) = p_i^{m_i} (1 - p_i)^{1 - m_i}
\end{equation}
where \( m_i \in \{0, 1\} \) indicates whether the token \( x_i \) is masked (\( m_i = 1 \)) or not (\( m_i = 0 \)).

The model is trained to predict the masked span by minimizing the \textit{cross-entropy loss}:
\begin{equation}
L_{\mathrm{mslm}} = - \frac{1}{\gamma} \sum_{i=1}^{\gamma} \log P(y_i | x_i)
\end{equation}
where $\gamma$ represents the maximum length of input sequence, and $y_i$ denotes the predicted output for each masked token $x_i$. The loss function trains the model to recover the masked spans, encouraging it to learn rich contextual information for span prediction.

\subsubsection{Span-level Contrastive Learning}

In addition to MSLM, we apply \textbf{Span-level Contrastive Learning} to further refine the traditional model’s ability to differentiate between different spans in the context of specific relations. Contrastive learning encourages the model to learn useful embeddings that can distinguish positive spans (relevant to the relation) from negative spans (irrelevant to the relation).

For each sentence $x$, the model learns to predict both the \textit{positive span} $s_p$ (e.g., ``was born in") and the \textit{negative span} $s_n$ (e.g., ``New York"). Let $h_a, h_p, h_n$ represent the embeddings of the anchor (relation type), positive, and negative spans, respectively. We optimize the following contrastive loss:

\begin{equation}
L_{\mathrm{scl}} = - \log \frac{\exp(\mathrm{sim}(h_a, h_p) / \tau)}{\exp(\mathrm{sim}(h_a, h_p) / \tau) + \exp(\mathrm{sim}(h_a, h_n) / \tau)}
\end{equation}

where \( \mathrm{sim}(h_a, h_x) \) is the cosine similarity between the anchor \( h_a \) and the span \( h_x \), and \( \tau \) is the temperature parameter controlling the sharpness of the similarity function.

This loss function encourages the model to bring positive spans closer to their respective anchor contexts and push negative spans further away, improving the model’s span-level understanding.

\subsubsection{Multi-task Training Objective}

The total loss for pre-training is a weighted combination of MSLM and SCL losses:

\begin{equation}
L_{\mathrm{pre-train}} = \lambda_1 L_{\mathrm{mslm}} + \lambda_2 L_{\mathrm{scl}},
\end{equation}
where \( \lambda_1 \) and \( \lambda_2 \) control the relative importance of the two tasks.

\subsection{Task-Oriented Fine-tuning}

After pre-training, the model is fine-tuned on the specific few-shot relation extraction task. Fine-tuning is performed using both the original small-shot examples and the synthetic samples generated in the previous stages. The goal of fine-tuning is to adapt the pre-trained model to the downstream task of relation classification, which involves predicting the correct relation between two entities in a given sentence.

For each input sentence $ x $ with entities \( e_{\mathrm{sub}} \) and \( e_{\mathrm{obj}} \) , the model is fine-tuned to predict the correct relation \( r \in \mathcal{R} \) from the set of relations. The input to the model consists of the sentence \( x_i \), along with the two entities \( e_{\mathrm{sub}_{i}} \) and \( e_{\mathrm{obj}_i} \). The output is the predicted relation label \( r_i \). The final relation prediction is made by passing the output of the encoder through a fully connected classification layer:

\begin{equation}
P(r_i | x_i, e_{\mathrm{sub}_{i}}, e_{\mathrm{obj}_i}; \theta) = \mathrm{softmax}(W \cdot h(x_i, e_{\mathrm{sub}_{i}}, e_{\mathrm{obj}_i}) + b)
\end{equation}

where \( h(x_i, e_{\mathrm{sub}_{i}}, e_{\mathrm{obj}_i}) \) is the representation of the sentence and entities from the pre-trained model, and \( W \) and \( b \) are the learned weights and bias of the classifier. The classification loss is computed as:

\begin{equation}
L_{\mathrm{finetune}} = - \sum_{i=1}^{N} \log P(r_i | x_i, e_{\mathrm{sub}_{i}}, e_{\mathrm{obj}_i}; \theta)
\end{equation}

where \( N \approx 2 \times K \times |\mathcal{R}|\) is the total number of samples in the training set, because the number of samples for some relation types may be less than $K$. And \( P(r_i | x_i, e_{\mathrm{sub}_{i}}, e_{\mathrm{obj}_i}; \theta) \) is the predicted probability for the correct relation \( r_i \).

\section{Experiments}

\subsection{Experimental Design}

\begin{table}[ht]
\small
\centering
\renewcommand{\arraystretch}{1.2} 
\resizebox{\columnwidth}{!}{
\begin{tabular}{cccccccc}
\hline
\multirow{2}{*}{Dataset} & \multicolumn{4}{c}{\#Train}         & \multirow{2}{*}{\#Val} & \multirow{2}{*}{\#Test} & \multirow{2}{*}{\#Rel} \\ \cline{2-5}
                         & 8-Shot & 16-Shot & 32-Shot & All    &                        &                         &                        \\ \hline
SemEval                  &   148     &   295      &    589     & 6,507  & 1,439                  & 2,717                   & 19                     \\
TACRED                   & 334    & 662     & 1,305   & 68,124 & 22,631                 & 15,509                  & 42                     \\
TACREV                   &  334      &   662      &   1,305      & 68,124 & 22,631                 & 15,509                  & 42                     \\
Re-TACRED                &     318   &    630     &    1,238     & 58,465 & 19,584                 & 13,418                  & 40                     \\ \hline
\end{tabular}
}
\caption{Statistics of our experimental datasets. K-shot means sampling K instances from each relation type. For relation types with fewer than K instances, we sample all available data. All refers to the complete training dataset. }
\label{table:1}
\end{table}

\paragraph{Datasets.} \textbf{SemEval 2010 Task 8} \cite{hendrickx2010semeval} is a widely used testbed for relation extraction. The schema is targeted at semantic relations between pairs of nominals, which requires a certain level of abstractive capabilities. \textbf{TACRED} \cite{zhang2017position} is a large-scale dataset annotated using Amazon Mechanical Turk crowdsourcing. It was initially created for the TAC knowledge base population and mainly covers common relations between people, organizations, and locations based on the TAC KBP scheme. \textbf{TACREV} \cite{alt2020tacred} is a label-corrected version of the TACRED dataset, which motivates from the unresolved challenging cases in the original TACRED dataset. \textbf{Re-TACRED} \cite{stoica2021re} further conducted a more comprehensive analysis and re-annotated the entire dataset. Besides, it made alternations to the schema to make it more clear and intuitive, which greatly improved the dataset quality. Following the strategy adopted by ~\cite{gao2021making,xu2023unleash}, for each relation type, we randomly sample K instances per relation (K-shot) for the training phase. The whole validation set and test set are preserved to ensure the effectiveness of the evaluation. The statistics of datasets are presented in \cref{table:1}.

\begin{table*}[ht]
    \centering
    \renewcommand{\arraystretch}{1.5} 
    \setlength{\tabcolsep}{2mm}
    \resizebox{\linewidth}{!}{
    \begin{tabular}{llccccccccccccc}
    \hline
    \multirow{2}{*}{\textbf{Paradigms}}                      & \multirow{2}{*}{\textbf{Models}} & \multicolumn{3}{c}{\textbf{SemEval}}          & \multicolumn{3}{c}{\textbf{TACRED}}           & \multicolumn{3}{c}{\textbf{TACREV}}           & \multicolumn{3}{c}{\textbf{Re-TACRED}}    &  \multirow{2}{*}{\textbf{Avg.}}     \\ \cline{3-14} 
                                                             &                                  & \textbf{k=8}  & \textbf{k=16} & \textbf{k=32} & \textbf{k=8}  & \textbf{k=16} & \textbf{k=32} & \textbf{k=8}  & \textbf{k=16} & \textbf{k=32} & \textbf{k=8}  & \textbf{k=16} & \textbf{k=32} \\ \hline
    \multirow{5}{*}{Traditional Methods} & CP       & 75.6          & 81.8          & 83.8          & 33.7          & 34.9          & 36.4          & 33.2          & 33.9          & 34.8          & 55.8          & 58.9          & 65.4  & 52.3\\
    & TYP Marker                       & 70.9          & 77.1          & 80.3          & 28.9          & 32.0          & 32.4          & 27.6          & 31.2          & 32.0          & 44.8          & 54.1          & 60.0       &   47.6\\
                                                             & KnowPrompt                       & 74.3          & 82.9          & 84.8          & 32.0          & 35.4          & 36.5          & 32.1          & 33.1          & 34.7          & 55.3          & 63.3          & 65.0         & 52.5\\
                                                             & GenPT                            & 77.1          & 81.5          & 83.9          & 35.7          & 36.6          & 37.4          & 34.4          & 34.6          & 36.2          & 57.2          & 60.4          & 65.2       &  53.3 \\
                                                             & PTRE                             & 79.0          & 83.0          & 84.6          & 32.7          & 36.5          & \underline{39.2}          & 33.9          & 36.3          & 38.1          & 58.8          & \underline{64.4}          & 66.3        & 54.4 \\ \hline
    \multirow{3}{*}{LLM-based Methods}   & LLama-2$^\star$                          & \multicolumn{3}{c}{56.9}                          & \multicolumn{3}{c}{26.5}                          & \multicolumn{3}{c}{27.3}                          & \multicolumn{3}{c}{38.1}                       &  37.2 \\
                                                             & GPT-3.5$^\star$                          & \multicolumn{3}{c}{60.5}                      & \multicolumn{3}{c}{29.7}                      & \multicolumn{3}{c}{30.0}                      & \multicolumn{3}{c}{39.6}                  & 39.9   \\
                                                             & GPT-4$^\star$                            & \multicolumn{3}{c}{65.3}                          & \multicolumn{3}{c}{32.3}                          & \multicolumn{3}{c}{32.1}                          & \multicolumn{3}{c}{45.9}                      &   43.9 \\ \hline
    \multirow{4}{*}{Hybrid Methods}      & Unleash (TYP Marker)                          & 78.3          & 81.7          & 84.8          & 35.8          & 36.6          & 37.1          & 36.7          & 36.5          & 37.0          & 58.4          & 60.6          & 64.8        &  54.0\\
                                                             & ${\mathrm{S}^{2}}$ynRE (CP)            & \underline{79.2}          & \underline{83.6}          & \underline{85.3}          & 36.3          & 37.2          & 38.7          & 35.2          & 36.1          & 36.9          & \underline{59.4}          & 63.9          & \underline{67.5}       &  54.9 \\
                                                            & \textbf{TKRE} (TYP Marker)                    & 77.6 & 81.7 & 84.1 & \underline{36.7} & \underline{38.3} & 39.1 & \underline{37.2} & \underline{38.2} & \underline{39.0} & 58.7 & 63.8       & 65.5      &   \underline{55.0} \\
                                                             & \textbf{TKRE} (GenPT)         & \textbf{80.7} & \textbf{84.9} & \textbf{86.6} & \textbf{39.6} & \textbf{42.1} & \textbf{43.2} & \textbf{40.2} & \textbf{41.3} & \textbf{42.9} & \textbf{61.3} & \textbf{67.6}          & \textbf{68.9}        & \textbf{58.3} \\   \hline
    \end{tabular}
    }
    \caption{Micro F1 (\%) of few-shot performance. The best results are in \textbf{bold} and the second best ones are \underline{underlined}. $^\star$ refers to the performance with one-shot demonstrations. ($\cdot$) refers to the backbone model used. }
    \label{table:main results}
    \vspace{-0mm}
\end{table*}

\paragraph{Baselines.} We compare our proposed TKRE with \textit{Traditional RE}, \textit{LLM-based} and \textit{Hybrid} Methods. The \textit{Traditional RE} methods include \textbf{CP} \cite{peng2020learning}, \textbf{TYP Marker} \cite{zhou2022improved}, \textbf{Knowprompt} \cite{chen2022knowprompt}, \textbf{GenPT} \cite{han2022generative}, and \textbf{PTRE} \cite{zhang2024prompt}. The \textit{LLM-based} methods include \textbf{LLaMA-2} \cite{touvron2023llama}, \textbf{GPT-3.5} \cite{ouyang2022training}, and \textbf{GPT-4} \cite{achiam2023gpt}. The \textit{Hybrid} methods include \textbf{Unleash} \cite{xu2023unleash}, and \textbf{$\mathbf{S}^{\mathbf{2}}$ynRE} \cite{xu2023s2ynre}. More details are shown in Appendix.

\paragraph{Implementation Details.} In the LLM data generation part, we leverage the 13B version for LLama-2 ( llama-2-13b-chat-hf ) to conduct experiments, and we double the K-shot training set through LLMs. For example, for an 8-shot training set, we construct 8 pieces of pseudo data per relation, thereby creating the final augmented training set.

In the Traditional RE Model Pre-Training part, we adopt TYP Marker and GenPT as the base architecture. Regarding the evaluation metrics, we adopt the micro F1 scores of RE as the primary metric to evaluate models, considering that F1 scores can assess the overall performance of precision and recall. More details are shown in Appendix.

\subsection{Main Results}
The main results are illustrated in Table \ref{table:main results}. Our proposed \textbf{TKRE} model outperforms all baselines across all metrics. Particularly compared to the base architecture TYP Marker and GenPT, our method manifests a significant advantage. This demonstrates the effectiveness of our designs and the benefits of integrating traditional RE models and LLMs. Furthermore, there are also some interesting phenomena:

1) The vast majority of methods exhibit superior performance on the ReTACRED dataset compared to the TACRED and TACREV datasets. This is reasonable as Re-TACRED is an improved version among these three datasets, which addresses some shortcomings of the original TACRED dataset and refactors its training/development/test sets. The more precise labels contribute to the learning process of these models, thereby yielding superior performance. 

2) Among these LLM-based methods, \textbf{GPT-4} demonstrates competitive performance and significantly outperforms \textbf{GPT-3.5} and \textbf{LLama-2} on all datasets. This proves its strong ability, as claimed in \cite{achiam2023gpt}. However, it exhibits the generation performance in FSRE. This may be because the training objective of \textbf{GPT-4} is on generative tasks, which predict the next word based on context, rather than optimizing specifically for RE tasks even though it utilized various very largescale corpora for training.

3) \textbf{Unleash} introduces a schema-constrained data augmentation method using LLMs to enhance the \textbf{TYP Marker} baseline by 6.4\%. Similarly, \textbf{$\mathbf{S}^{\mathbf{2}}$ynRE}
presents a two-stage self-training framework leveraging LLM-generated synthetic data, improving the \textbf{CP} baseline by 2.6\%. Both approaches demonstrate a measurable improvement over their respective baselines, validating the feasibility of this line of research. In comparison, our \textbf{TKRE} model significantly outperforms \textbf{TYP Marker} by 7.8\% and \textbf{GenPT} by 5.0\%, further confirming the effectiveness of our approach from a different perspective.

\subsection{Ablation Study}

To validate the effectiveness of main components in TKRE, we introduce the following variants for the ablation study:

\textbf{TKRE} \textit{w/o} EDKG: This variant removes explanation-driven knowledge generation at the data preparation stage.

\textbf{TKRE} \textit{w/o} SCDG: This variant removes schema-constrained data generation at the data preparation stage.

\textbf{TKRE} \textit{w/o} MSLM: This variant abides by the mask setting of BERT\cite{devlin2019bert} instead of masked span language modeling at the two-stage knowledge guidance pre-Training stage.

\textbf{TKRE} \textit{w/o} SCL: This variant skips the span-level contrastive learning at the two-stage knowledge guidance pre-Training stage.

\textbf{TKRE} \textit{w/o ALL}: This variant addresses the FSRE task by fine-tuning the baseline model (e.g., TYP Marker) solely on the golden training dataset, without incorporating any of the enhancements introduced by our method.

\begin{table}[htb]
    \centering
    \footnotesize 
    \setlength{\tabcolsep}{0.8mm}
    \resizebox{\linewidth}{!}{
    \begin{tabular}{lccccccc}
    \toprule
        \multirow{2}{*}{\textbf{Models}}  & \multicolumn{3}{c}{\textbf{SemEval}} & \multicolumn{3}{c}{\textbf{TACRED}} & \multirow{2}{*}{\textbf{Avg.}}\\
        \cmidrule(lr){2-4} \cmidrule(lr){5-7}
        & {\textbf{k=8}} & {\textbf{k=16}}  & {\textbf{k=32}} & {\textbf{k=8}} & {\textbf{k=16}}  & {\textbf{k=32}} \\
        \cmidrule(lr){1-8} 
        \textbf{TKRE} & \textbf{77.6} & \textbf{81.7} & \textbf{84.1} & \textbf{36.7} & \textbf{38.3} & \textbf{39.1} & \textbf{59.6} \\
        \cmidrule(lr){1-8}   
        \textit{w/o} EDKG & 73.1 & 75.8 & 80.2 & 30.9 & 33.7 & 36.1 &  55.0\\
        \textit{w/o} SCDG & 75.2 & 79.3 & 82.1 & 34.1 & 35.3 & 37.3 &  57.2\\
        \textit{w/o} MSLM & 75.3 & 78.5 & 81.1 & 35.4 & 36.3 & 36.9 &  57.2\\
        \textit{w/o} SCL & 76.2 & 79.8 & 82.9 & 35.2 & 35.9 & 37.4 &  57.9 \\
        \textit{w/o ALL} & 70.9 & 77.1 & 80.3 & 28.9 & 32.0 & 32.4 &  53.6 \\
    \bottomrule
    \end{tabular}
    }
    \vspace{-0.2cm}
    \caption{Ablation study results for TACRED.}
    \label{tab:ablation}
    \vspace{-0.3cm}
\end{table}

From Table \ref{tab:ablation}, we observe the following:

1) Removing EDKG significantly reduces performance, especially in smaller settings, highlighting its importance in generating relation-relevant explanations for FSRE.

2) The absence of SCDG leads to a performance decline, especially on TACRED with fewer training examples. This demonstrates the importance of schema constraints help LLMs toward more accurate and relevant data generation.

3) Replacing MSLM with traditional BERT-style masking results in a modest performance reduction, emphasizing the significance of span-based masking in capturing relational dependencies.

4) Omitting SCL leads to a slight drop in performance, suggesting that it plays a key role in differentiating relevant spans, which is crucial for fine-grained RE.

5) The variant with all enhancements removed exhibits the lowest performance, particularly for smaller K values. This reinforces the necessity of these enhancements in boosting the performance of FSRE.

\subsection{Analysis and Discussions}
\vspace{-0.2cm}
\begin{figure}[h]
    \small
	\centering
	\includegraphics[width=0.99\linewidth]{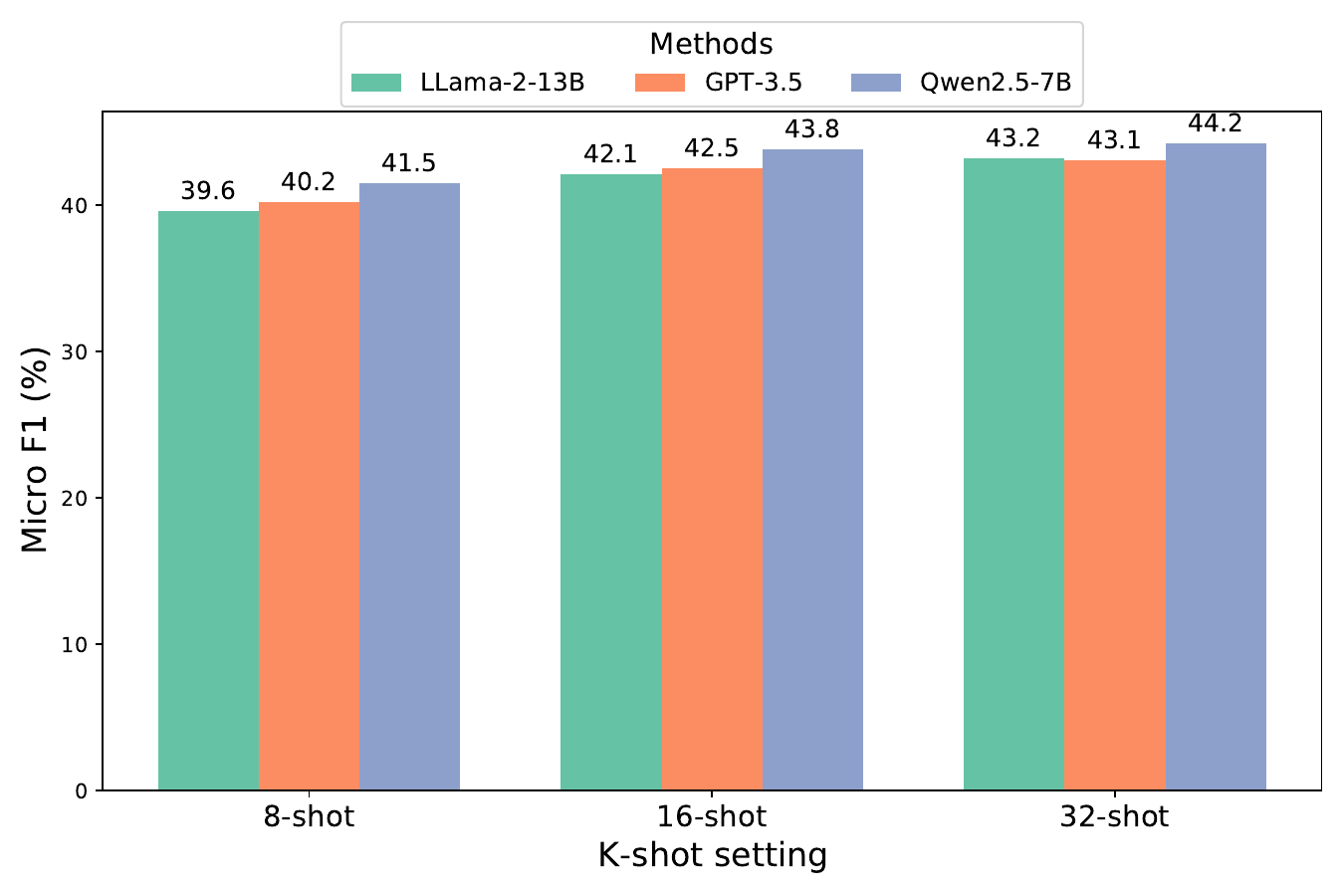}
        \vspace{-0.4cm}
        \caption{Micro F1 (\%) of \textbf{TKRE} (GenPT) with generated explanation-driven knowledge by various LLMs.}
	\label{exp1}
\end{figure}
\vspace{-0.5cm}
\paragraph{Explanation-driven Knowledge Generated from Different LLMs.} We evaluate the impact of different LLMs applied to generate explanation-driven knowledge. We extend \textbf{GPT-3.5} and \textbf{Qwen2.5-7B} \cite{team2024qwen2} as other knowledge generators under the TACRED dataset, and continue model pre-training as well as fine-tuning under the same setting of Llama-2. As shown in Figure \ref{exp1}, the models pre-trained on different LLMs have similar performance. This indicates that our framework is not sensitive to different LLMs for FSRE. 

\begin{figure}[h]
    \small
	\centering
	\includegraphics[width=0.99\linewidth]{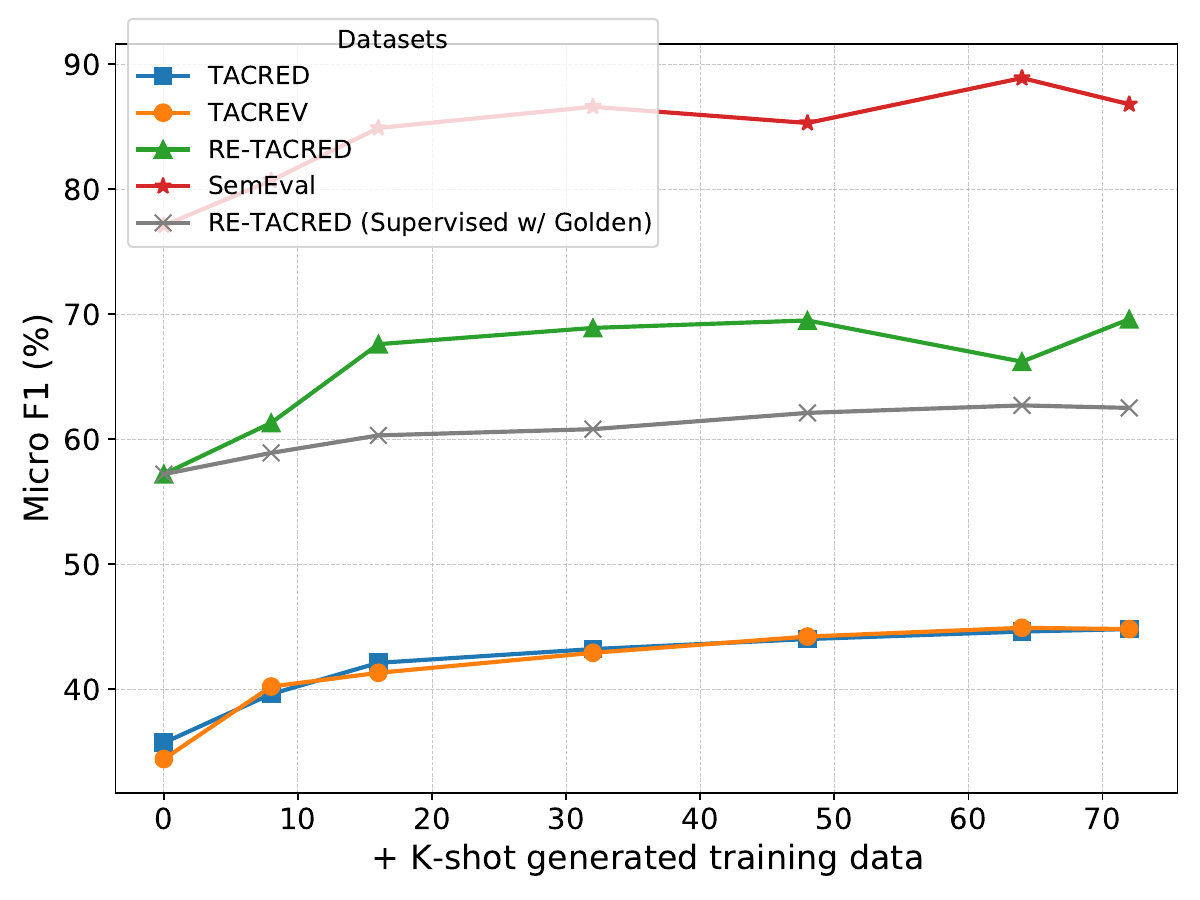}
        \vspace{-0.4cm}
        \caption{Micro F1 (\%) of \textbf{TKRE} (GenPT) with generated training data and original 8-shot data.}
	\label{exp2}
        \vspace{-0.3cm}
\end{figure}
\paragraph{Scale of Synthetic data.} As shown in Figure \ref{exp2}, the performance improves initially with the increasing volume of schema-constrained data but eventually declines, while still consistently surpassing the performance of the scenario without generated data. This observation suggests that incorporating schema-constrained data generated by LLMs can enhance FSRE performance up to a certain threshold. However, beyond this point, the inclusion of low-quality or noisy data begins to negatively affect the training process, as further discussed in the Appendix. This behavior is consistent with previous studies indicating that excessive noise in training data can undermine model performance, even considering the noise-resilience capabilities of LLMs~\cite{song2022learning}. 

We also report supervised results using additional golden training data to measure the utility of synthetic data. We can achieve two conclusions: 1) the advantage of golden training data are more significant when it is scaled up. However, this also takes more expensive costs. 2) TKRE demonstrates the ability to achieve performance on par with models trained using significantly larger volumes of manually annotated golden data, despite being trained with only a small amount of synthetic data. Furthermore, the generation of the requisite synthetic data requires only a few minutes of GPU computation.


\section{Conclusion}

In this work, we propose TKRE, a Two-Stage Knowledge Guided framework for FSRE, which effectively bridges the gap between generative and discriminative learning paradigms. By leveraging LLMs for explanation-driven knowledge generation and schema-constrained synthetic data, TKRE addresses key challenges in FSRE, including data scarcity and limited generalization ability. Our innovative two-stage knowledge-guided pre-training approach, which integrates Masked Span Language Modeling and Span-Level Contrastive Learning, enhances traditional FSRE models' ability to capture relational reasoning and fine-grained semantic distinctions.
Extensive experiments on multiple benchmark datasets demonstrate that TKRE significantly improves the performance of traditional FSRE models. Ablation studies further validate the effectiveness of each proposed component.
Future work could explore extending TKRE to broader relation extraction tasks, such as zero-shot scenarios, or further optimizing the generation process to mitigate noise in synthetic data.

\bibliographystyle{named}
\bibliography{ijcai24}

\clearpage

\appendix
\section{Experimental Details}
\label{appendix_exp}

\subsection{Baselines}
1) Traditional RE methods:
\begin{itemize}
    \item \textbf{CP} \cite{peng2020learning} proposes a contrastive learning pretext task that encourages sentence representations with the same relation to be similar and different ones to be disparate.
    
    \item \textbf{TYP Marker} \cite{zhou2022improved} proposes to incorporate entity representations with typed markers, which presents remarkable performance on the RE task.
    
    \item \textbf{Knowprompt} \cite{chen2022knowprompt} proposes a novel generative prompt tuning method to reformulate relation classification as an infilling problem, which exploits rich semantics of entity and relation types.
    
    \item \textbf{GenPT} \cite{han2022generative} proposes a novel generative prompt tuning method to reformulate relation classification as an infilling problem, which exploits rich semantics of entity and relation types.

    \item \textbf{PTRE} \cite{zhang2024prompt} proposes a prompt-tuning method that incorporates global and local relation graphs, along with semantic knowledge from relation labels, to improve few-shot relation extraction. Experiments on four datasets show its effectiveness in low-resource settings.
\end{itemize}

2) LLM-based Methods:
\begin{itemize}
    \item \textbf{LLaMA-2} \cite{touvron2023llama}, \textbf{GPT-3.5} \cite{ouyang2022training}, and \textbf{GPT-4} \cite{achiam2023gpt} are advanced LLMs, with GPT-4 particularly standing out as the state-of-the-art (SOTA) model among them. In our study, we utilize the GPT-3.5 API and directly instruct the GPT-4 model (with 1.8 trillion parameters) provided by OpenAI. Additionally, we adopt the 13B parameter version of LLaMA-2 (llama-2-13b-chat-hf). The results are obtained by directly instructing these LLMs using the same in-context learning prompt as described in \cite{xu2023unleash}.
\end{itemize}

3) Hybrid Methods:
\begin{itemize}
    \item \textbf{Unleash} \cite{xu2023unleash} proposes a novel generative prompt tuning method to reformulate relation classification as an infilling problem, which exploits rich semantics of entity and relation types.
    
    \item \textbf{${\mathrm{S}^{2}}$ynRE} \cite{xu2023s2ynre} proposes a two-stage self-training framework that uses large language models to generate synthetic data, improving relation extraction by alternating between synthetic and golden data. Experiments show its effectiveness.
\end{itemize}
\subsection{Implementation Details}
The approach is based on Pytorch and the Transformer library of Huggingface. We utilize 4 NVIDIA Tesla A800 GPU with 80 GB memory to run all experiments. For a fair comparison, in the Traditional RE Model Pre-Training part, we adopt RoBERTa-large ~\cite{liu2019roberta} as the base architecture for our implementation based on both TYP Marker and GenPT, and the model is optimized by AdamW ~\cite{loshchilov2017decoupled} with a learning rate of $3e{-5}$. We use a batch size of 32, which are chosen for practical consideration to fit into GPU memory. We train the model on the training set for 10 epochs and choose the best epoch based on the micro-F1 performance on the development set.
\section{Case Analysis}
\label{appendix_case}

\subsection{Wrong Cases from GenPT and GPT-4}
\label{app:wrong}
\begin{table*}[htbp]
  \centering
  \scalebox{0.74}{
    \begin{tabular}{cp{22.4em}llll}
    \toprule
    Dataset & \multicolumn{1}{c}{Case} & \multicolumn{1}{c}{Golden Relation} & \multicolumn{1}{c}{GenPT} & \multicolumn{1}{c}{GPT-4} & \multicolumn{1}{c}{TKRE(ours)}\\
    \midrule
    \multirow{1}[4]{*}{SemEval} & Context: The U.S. Transportation Security Administration (TSA) has lost a hard drive containing information about 100,000 former and current employees. \newline{}Head Type: -. Head Entity: information.\newline{}Tail Type: -. Tail Entity: employees & Message-Topic & Content-Container \ding{55} & Content-Container \ding{55} & \ding{52}\\
    \midrule
    \multirow{1}[4]{*}{TACRED} & Context:With his perfect English and British education (a photo on the wall of his study shows him as a teenager on the rugby team of Malvern College), Rashid became what he calls the “intellectual repository” for Western journalists who parachuted into the Afghan capital for the Soviet Union's last big invasion.\newline{}Head Type: PERSON. Head Entity: he. \newline{}Tail Type: MISC. Tail Entity: Western. & no\_relation & per:affiliation \ding{55}& per:employee\_of \ding{55}& \ding{52}\\
    \midrule
    \multirow{1}[4]{*}{TACREV} & Context: Following is a list of 'village idiots' bound and gagged in the straitjackets of my KILLFILE: Nickname: (optional), Firnando, Sabastawi, LeNoir, FARIS JAWAD, ABU ALWAFA, Franko Pizza, Katrina, Mujahid, Nasser, Sinbad, Saif Al-Islam, Free Palestine, Salem, Nidal, Si-Salah, Troll Hunter, R Geovani, Rudolph W. Giuliani, Ilan Ramon, Blondes Gaulloises, Romero, Salim.\newline{}Head Type: PERSON. Head Entity: Ilan Ramon.\newline{}Tail Type: LOCATION. Tail Entity: Katrina. & no\_relation & per:family \ding{55}& per:acquaintance \ding{55} & \ding{52}\\
    \midrule
    \multirow{1}[4]{*}{RE-TACRED} & Context: Ms. Herold had suffered a series of heartbreaking losses over the last several years, beginning with the death of her only child, then her husband, then her beloved chimp Travis, as well as the tragic maiming of friend and employee Charla Nash,” Golger said in a statement.\newline{} Head Type: PERSON. Head Entity: her.\newline{}Tail Type: PERSON. Tail Entity: Charla Nash. & no\_relation & per:employee\_of \ding{55}&  per:employee\_of \ding{55}& \ding{52} \\
    \bottomrule
    \end{tabular}}
    \caption{Wrong cases predicted by both GenPT and GPT-4. The golden relation categories are listed in the third column.}
  \label{tab:wrong}
\end{table*}

From Table \ref{tab:wrong}, we notice that some RE instances are challenging for Traditional methods and LLM-based methods: 1) LLMs exhibit poor sensitivity to entity order and long-range dependencies. For instance, in the SemEval case, GPT-4 misclassifies the "information-employees" relationship as Content-container instead of Message-Topic. This error arises because LLMs often conflate semantically adjacent relations when entity mentions span distant contexts or lack explicit positional cues. 2) Traditional methods are prone to confusion from overlapping entity types or ambiguous labels. In the TACRED example, both GPT-4 and GenPT incorrectly infer employment relations (e.g., per:employee\_of ) between "he" (PERSON) and "Western" (MISC), despite no explicit evidence. This also reflects LLMs’ tendency to overfit surface patterns while neglecting fine-grained type constraints. 3) LLMs treat RE as a text generation task, leading to misalignment with discriminative objectives. The TACREV case illustrates this: GPT-4 hallucinates per:acquaintance between "Ilan Ramon" and "Katrina" (LOCATION), likely due to spurious correlations in training data, whereas TKRE avoids such noise by integrating structured reasoning. However, our proposed TKRE can address these difficult cases very well, which again demonstrates the effectiveness of our designs and the benefits of integrating traditional RE models and LLMs.

\subsection{Generated explanation-driven knowledge from LLMs}
There are some cases for generated explanation from GPT-3.5 in ~\cref{tab:DATA1}. Through careful observation of the generated explanations from GPT-3.5, we find that LLMs often exhibit a strong capacity for logical reasoning and analysis. When explaining the relationships already labeled in the samples, the models demonstrate coherent justifications that reflect an understanding of the underlying patterns. This suggests that these LLMs not only retrieve factual information but also possess the ability to reason about the relations between entities, providing valuable insights into the model’s interpretability and knowledge generation process.

\begin{table*}[htbp]
  \centering
  \scalebox{0.74}{
    \begin{tabular}{cp{22.4em}lp{22.4em}}
    \toprule
    Dataset & \multicolumn{1}{c}{Case} & \multicolumn{1}{c}{Golden Relation} & \multicolumn{1}{c}{Explanation-driven Knowledge} \\
    \midrule
    \multirow{1}[4]{*}{SemEval} & Context: The U.S. Transportation Security Administration (TSA) has lost a hard drive containing information about 100,000 former and current employees. \newline{}Head Type: -. Head Entity: information.\newline{}Tail Type: -. Tail Entity: employees & Message-Topic & In this sentence, the subject "information" has the relation "Message-Topic" with the object "employees" because the "information" in question pertains to or is about the "employees." Specifically, the lost hard drive contains details about 100,000 former and current employees. The "information" acts as the message, and the "employees" are the topic or subject matter of that message. In terms of the Message-Topic relationship: Message: The content being conveyed or lost, which in this case is "information." Topic: The subject or focus of the message, which is "employees." So, the "information" on the hard drive revolves around the "employees," making the employees the topic of the information in question. \\
    \midrule
    \multirow{1}[4]{*}{TACRED} & Context: With his perfect English and British education (a photo on the wall of his study shows him as a teenager on the rugby team of Malvern College), Rashid became what he calls the “intellectual repository” for Western journalists who parachuted into the Afghan capital for the Soviet Union's last big invasion.\newline{}Head Type: PERSON. Head Entity: he. \newline{}Tail Type: MISC. Tail Entity: Western. & no\_relation &  In the sentence provided, the subject "he" (referring to Rashid) has the relation "no\_relation" with the object "Western" because "he" is not directly connected or related to "Western" in any specific way that implies a typical syntactic or semantic relationship. Here’s why: "Western" in this context is used as an adjective to describe the journalists, not as a noun or concept that is directly linked to "he" (Rashid). Rashid is described as an intellectual figure for "Western journalists," but there’s no explicit relationship or action involving "he" and "Western." He is not a part of "Western" in any direct sense, nor is he interacting with it in a way that would define a relationship like "is\_a," "has," or similar. Thus, "he" (Rashid) and "Western" (as a descriptor of the journalists) do not form a meaningful connection in terms of subject-object relations, so the appropriate relation is "no\_relation."\\
    \midrule
    \multirow{1}[4]{*}{TACREV} &  Context: Following is a list of 'village idiots' bound and gagged in the straitjackets of my KILLFILE: Nickname: (optional), Firnando, Sabastawi, LeNoir, FARIS JAWAD, ABU ALWAFA, Franko Pizza, Katrina, Mujahid, Nasser, Sinbad, Saif Al-Islam, Free Palestine, Salem, Nidal, Si-Salah, Troll Hunter, R Geovani, Rudolph W. Giuliani, Ilan Ramon, Blondes Gaulloises, Romero, Salim.\newline{}Head Type: PERSON. Head Entity: Ilan Ramon.\newline{}Tail Type: LOCATION. Tail Entity: Katrina. & no\_relation & In this sentence, the subject "Ilan Ramon" has the relation "no\_relation" with the object "Katrina" because there is no direct or meaningful connection established between these two entities in the context of the sentence. Here's the breakdown: "Ilan Ramon" refers to a person, likely the Israeli astronaut. "Katrina" is presumably referring to a name, but there's no additional context to link "Katrina" to "Ilan Ramon" in terms of the sentence's structure or meaning. Both "Ilan Ramon" and "Katrina" are listed as part of the same "KILLFILE" (a list of people or things the speaker wishes to ignore or block), but there is no interaction or relationship implied between them. The sentence simply lists various names, with no semantic or syntactic connection between "Ilan Ramon" and "Katrina." Thus, the appropriate relation between "Ilan Ramon" and "Katrina" is "no\_relation" because they are simply two separate items in the list with no inherent relationship. \\
    \midrule
    \multirow{1}[4]{*}{RE-TACRED} & Context: Ms. Herold had suffered a series of heartbreaking losses over the last several years, beginning with the death of her only child, then her husband, then her beloved chimp Travis, as well as the tragic maiming of friend and employee Charla Nash,” Golger said in a statement.\newline{} Head Type: PERSON. Head Entity: her.\newline{}Tail Type: PERSON. Tail Entity: Charla Nash. & no\_relation &  In this sentence, the subject "her" (referring to Ms. Herold) has the relation "no\_relation" with the object "Charla Nash" because there is no direct or meaningful relationship implied between "her" and "Charla Nash" in terms of subject-object syntax. Here’s why: "Her" refers to Ms. Herold, the subject of the sentence who has experienced various losses. "Charla Nash" is mentioned as a person who was tragically maimed, and while Charla Nash is part of the events Ms. Herold has gone through, she is not in a direct relationship with the subject "her" in a way that implies an action, possession, or descriptive link. "Charla Nash" is an individual who suffered a tragic event in relation to Ms. Herold’s story, but there is no direct interaction between Ms. Herold and Charla Nash described here (e.g., no "is\_a," "has," or "part\_of" relationship). Thus, while Charla Nash is part of the tragic events described, "her" (Ms. Herold) does not have a syntactic or semantic relationship with "Charla Nash" in this context. Therefore, the appropriate relation is "no\_relation."\\
    \bottomrule
    \end{tabular}}
    \caption{Generated explanation-driven knowledge from LLMs.}
  \label{tab:DATA1}
\end{table*}
\subsection{Schema-Constrained Data Generated from LLMs}
There are some cases for synthetic data from LLama-2 and GPT-3.5 in ~\cref{tab:DATA2}. Through human checks on 100 generated samples per dataset, we observe that labeled data generated by LLama-2 and GPT-3.5 are mostly correct. We think better performance results from GPT-3.5’s help. However, we also find that Some generated data from GPT-3.5 are of less quality than golden data. Through human checks, some generated samples are too short and concatenated by some semantically irrelevant sentences.
\begin{table*}[htbp]
  \centering
  \scalebox{0.74}{
    \begin{tabular}{cp{22.4em}p{22.4em}}
    \toprule
    Relation Type & \multicolumn{1}{c}{Case Generated by LLama-2} & \multicolumn{1}{c}{Case Generated by GPT-3.5} \\
    \midrule
    \multirow{1}[4]{*}{per:origin} & Context: The company 's stock price plummeted after it was revealed that the CEO had been embezzling funds for personal gain. \newline{}Head Type: PERSON. Head Entity: CEO.\newline{}Tail Type: ORGANIZATION. Tail Entity: company & Context: The stock price of the company dropped sharply after the CEO was found guilty of embezzlement. \newline{}Head Type: PERSON. Head Entity: CEO.\newline{}Tail Type: ORGANIZATION. Tail Entity: company \\
    \midrule
    \multirow{1}[4]{*}{org:stateorprovince\_of\_headquarters} & Context: The headquarters of the National Football League is located in New York City.\newline{}Head Type: ORGANIZATION. Head Entity: National Football League. \newline{}Tail Type: LOCATION. Tail Entity: New York City. &  Context: The National Football League's headquarters is based in New York City. \newline{}Head Type: ORGANIZATION. Head Entity: National Football League. \newline{}Tail Type: LOCATION. Tail Entity: New York City\\
    \midrule
    \multirow{1}[4]{*}{per:age} &  Context: The 92-year-old actress has won numerous awards for her performances on stage and screen.\newline{}Head Type: PERSON. Head Entity: actress.\newline{}Tail Type: DURATION. Tail Entity: 92-year-old. & Context: The actress, aged 54, has received multiple awards throughout her career. \newline{}Head Type: PERSON. Head Entity: actress. \newline{}Tail Type: DURATION. Tail Entity: 54-year-old \\
    \midrule
    \multirow{1}[4]{*}{org:subsidiaries} & Context: TOKYO 2007-04-13 04:03:53 UTC Nissan tying up with NEC in batteries for ecological vehicles NEC Tokin Corp , an NEC subsidiary that produces lithium-ion batteries , will handle the production , according to The Nikkei.\newline{} Head Type: ORGANIZATION. Head Entity: NEC.\newline{}Tail Type: ORGANIZATION. Tail Entity: NEC Tokin Corp. &  Context: NEC Tokin Corp, a subsidiary of NEC, will produce lithium-ion batteries for Nissan's ecological vehicles. \newline{}Head Type: ORGANIZATION. Head Entity: NEC.\newline{}Tail Type: ORGANIZATION. Tail Entity: NEC Tokin Corp\\
    \midrule
    \multirow{1}[4]{*}{org:founded\_by} & Context: The non-profit organization was founded by Jane Smith in 2005.\newline{} Head Type: ORGANIZATION. Head Entity: The non-profit organization.\newline{}Tail Type: PERSON. Tail Entity: Jane Smith. &  Context: Jane Smith established the non-profit organization in 2005. \newline{} Head Type: ORGANIZATION. Head Entity: The non-profit organization.\newline{}Tail Type: PERSON. Tail Entity: Jane Smith\\
    \midrule
    \multirow{1}[4]{*}{per:employee\_of} & Context: He served as a software engineer at Google before joining Amazon as a senior software developer.\newline{} Head Type: PERSON. Head Entity: He.\newline{}Tail Type: ORGANIZATION. Tail Entity: Google. &  Context: He worked at Google as a software engineer and later moved to Amazon as a senior software developer. \newline{} Head Type: PERSON. Head Entity: He.\newline{}Tail Type: ORGANIZATION. Tail Entity: Google\\
    \midrule
    \multirow{1}[4]{*}{per:title} & Context: The new policy , which was announced by ABC president Steve McPherson , will go into effect on August 1.\newline{} Head Type: PERSON. Head Entity: Steve McPherson.\newline{}Tail Type: TITLE. Tail Entity: president. &  Context: The president of ABC, Steve McPherson, announced the new policy which takes effect on August 1. \newline{} Head Type: PERSON. Head Entity: Steve McPherson.\newline{}Tail Type: TITLE. Tail Entity: president\\
    \midrule
    \multirow{1}[4]{*}{org:top\_members/employees} & Context: ``We're looking for someone who can help us expand our reach and engagement on social media." said Emily Chen, social media manager at Nike.\newline{} Head Type: ORGANIZATION. Head Entity: Nike.\newline{}Tail Type: PERSON. Tail Entity: Emily Chen. &  Context: Emily Chen, social media manager at Nike, stated they are looking for someone to help boost their social media presence. \newline{} Head Type: ORGANIZATION. Head Entity: Nike.\newline{}Tail Type: PERSON. Tail Entity: Emily Che\\
    \midrule
    \multirow{1}[4]{*}{per:stateorprovince\_of\_birth} & Context: The actor was born in Los Angeles , Calif. , on July 8 , 1968.\newline{} Head Type: PERSON. Head Entity: actor.\newline{}Tail Type: PERSON. Tail Entity: Calif. &  Context: The actor was born on July 8, 1968, in Los Angeles, California. \newline{} Head Type: PERSON. Head Entity: actor.\newline{}Tail Type: PERSON. Tail Entity: Calif.\\
    \midrule
    \multirow{1}[4]{*}{per:countries\_of\_residence} & Context: The US has a long history of intervention in Latin American countries including Cuba , Chile and Nicaragua.\newline{} Head Type: COUNTRY. Head Entity: US.\newline{}Tail Type: COUNTRY. Tail Entity: Cuba. &  Context: The United States has been involved in various Latin American countries such as Cuba, Chile, and Nicaragua throughout history. \newline{} Head Type: COUNTRY. Head Entity: US.\newline{}Tail Type: COUNTRY. Tail Entity: Cuba\\
    \midrule
    \multirow{1}[4]{*}{per:country\_of\_death} & Context: Former US Senator Ted Stevens dies in plane crash in Alaska.\newline{} Head Type: PERSON. Head Entity: Ted Stevens.\newline{}Tail Type: COUNTRY. Tail Entity: Alaska. &  Context: Ted Stevens, a former US Senator, passed away in a plane crash in Alaska. \newline{} Head Type: PERSON. Head Entity: Ted Stevens.\newline{}Tail Type: COUNTRY. Tail Entity: Alaska.\\
    \midrule
    \multirow{1}[4]{*}{org:founded} & Context: The World Health Organization was established in 1948 to promote global health and prevent disease.\newline{} Head Type: ORGANIZATION. Head Entity: World Health Organization.\newline{}Tail Type: DATE. Tail Entity: 1948. &  Context: The World Health Organization (WHO) was founded in 1948 to improve public health globally. \newline{} Head Type: ORGANIZATION. Head Entity: World Health Organization.\newline{}Tail Type: DATE. Tail Entity: 1948\\
    \midrule
    \multirow{1}[4]{*}{per:charges} & Context: A former Enron executive, Andrew Fastow , has been charged with money laundering and securities fraud.\newline{} Head Type: PERSON. Head Entity: Andrew Fastow.\newline{}Tail Type: CRIMINAL\_CHARGE. Tail Entity: money laundering. &  Context: Andrew Fastow, a former executive at Enron, has been charged with securities fraud and money laundering. \newline{} Head Type: PERSON. Head Entity: Andrew Fastow.\newline{}Tail Type: CRIMINAL\_CHARGE. Tail Entity: money laundering\\
    \bottomrule
    \end{tabular}}
    \caption{Parts of synthetic data from LLama-2 and GPT-3.5.}
  \label{tab:DATA2}
\end{table*}
\end{document}